\documentclass[conference]{IEEEtran}
\IEEEoverridecommandlockouts
\usepackage{cite}
\usepackage{amsmath,amssymb,amsfonts}
\usepackage{algorithmic}
\usepackage{graphicx}
\usepackage{textcomp}
\usepackage{xcolor}
\usepackage{multirow}
\usepackage{booktabs}
\usepackage{subcaption}
\usepackage{comment}

\usepackage[left=0.75in, right=0.75in, top=0.75in, bottom=0.75in]{geometry}
\def\BibTeX{{\rm B\kern-.05em{\sc i\kern-.025em b}\kern-.08em
    T\kern-.1667em\lower.7ex\hbox{E}\kern-.125emX}}

\makeatletter
\newcommand{\linebreakand}{%
  \end{@IEEEauthorhalign}
  \hfill\mbox{}\par
  \mbox{}\hfill\begin{@IEEEauthorhalign}
}
\makeatother
    
\begin{document}

\title{REPAIR-Bench: A Benchmark for Robot Error Perception And Interaction Recovery
}

\author{\IEEEauthorblockN{1\textsuperscript{st} Giuliano Pioldi}
\IEEEauthorblockA{
% \textit{University Name} \\
\textit{Cornell University} \\
\textit{Cornell Tech} \\
New York, NY \\
gp433@cornell.edu}
%email@cornell.edu}
\and
\IEEEauthorblockN{2\textsuperscript{nd} Yashika Batra}
\IEEEauthorblockA{
% \textit{University Name} \\
\textit{Cornell University} \\
\textit{Cornell Tech} \\
New York, NY \\
yb344@cornell.edu}
%email@cornell.edu}
\and
\IEEEauthorblockN{3\textsuperscript{rd} Arman Ibrayeva}
\IEEEauthorblockA{
% \textit{University Name} \\
\textit{Cornell University} \\
\textit{Cornell Tech} \\
New York, NY \\
ai342@cornell.edu}
%email@cornell.edu}
\and
\IEEEauthorblockN{4\textsuperscript{th} Yuanchen Bai}
\IEEEauthorblockA{
% \textit{University Name} \\
\textit{Cornell University} \\
\textit{Cornell Tech} \\
New York, NY \\
yb299@cornell.edu}
%email@cornell.edu}

\linebreakand

\IEEEauthorblockN{5\textsuperscript{th} Purnjay Maruur}
\IEEEauthorblockA{
% \textit{University Name} \\
\textit{Cornell University} \\
\textit{Cornell Tech} \\
New York, NY \\
pm674@cornell.edu}
%email@cornell.edu}
\and
\IEEEauthorblockN{6\textsuperscript{th} Promise Ekpo}
\IEEEauthorblockA{
% \textit{University Name} \\
\textit{Cornell University} \\
\textit{Cornell Tech} \\
New York, NY \\
poe6@cornell.edu}
%email@cornell.edu}
\and
\IEEEauthorblockN{7\textsuperscript{th} Angelique Taylor}
\IEEEauthorblockA{
% \textit{University Name} \\
\textit{Cornell University} \\
\textit{Cornell Tech} \\
New York, NY \\
amt298@cornell.edu}
%email@cornell.edu}
}

\newgeometry{top=1in, bottom=0.75in, left=0.75in, right=0.75in}
\maketitle

% \begin{abstract}
% Robots deployed in high-stakes settings frequently encounter failures due to perception errors, time constraints, miscomprehension, or environmental uncertainty. Characterizing how failures unfold and how users perceive and respond to them is essential for building robots that maintain user trust and task effectiveness. We introduce REPAIR-Bench, a benchmark for embodied failure understanding that integrates failure progression, multimodal user behavior, and user perception. Built upon the Human-Robot Interaction (RFM-HRI) dataset of 214 interaction samples from 41 participants, the benchmark supports three tasks across two dimensions: sliding-window failure classification, temporal failure detection, and recovery preference prediction \cite{rfm-hri}. Together, these tasks enable systematic analysis of failure progression and user adaptation, and analyze how different modalities contribute to failure dynamics.

\begin{abstract}
%Robots deployed in high-pressure settings inevitably encounter failures due to perception errors, time constraints, miscomprehension, or environmental uncertainty. Characterizing how failures unfold and how users perceive and respond to them is essential for building robots that maintain user trust and task effectiveness, yet prior work often treats failures in isolation and emphasizes binary detection over temporal and user-centered modeling. We introduce Robot Error Perception And Interaction Recovery-Bench (REPAIR-Bench), a benchmark for embodied failure understanding that integrates failure progression, multimodal behavioral signals, and user recovery preferences. Built on 214 interaction sessions from 41 participants in the Robot Failure in Medical Human Robot Interaction (RFM-HRI) dataset, the benchmark spans four induced failure types and includes synchronized facial action units, head pose, speech transcripts, and post-interaction affect and recovery reports. By formalizing failure as a temporally evolving, multimodal, and user-centered phenomenon, REPAIR-Bench provides the HRI community with a standardized framework for evaluating embodied failure understanding and advancing more transparent, adaptive, and trustworthy robot systems for safety-critical applications such as in healthcare settings.

Understanding how users perceive and respond to robot failures is essential for building robust and trustworthy robot systems. Prior work, however, (i) often treats failures as independent events,  (ii) emphasizes binary failure detection, (iii) with rule-based recovery modeling. We present REPAIR-Bench, built on 214 interaction trials from 41 participants, the benchmark spans four induced failure types and provides synchronized facial action units, head pose, speech transcripts, and post-interaction affect and recovery reports. The benchmark spans three novel evaluation tasks that jointly capture the lifecycle of failure in human-robot interaction (HRI): (i) failure detection over inter-dependent interaction sessions, modeling longitudinal user adaptation across repeated failures; (ii) visual failure-type classification beyond binary success/failure formulations; and (iii) user-centered recovery prediction, inferring users' preferred recovery strategies from interaction context rather than relying on manually designed or rule-based strategies.
In baseline experiments, hierarchical recurrent modeling improved failure detection over a single-session model (strict F1: $0.80$ vs.\ $0.68$), achieved a failure localization mean signed error of $-0.51$\,s, median absolute error of 2.97 s and, for recovery prediction, a QLoRA-tuned Mistral-7B reached Hit@$5=0.76$ and F1@$5=0.32$. 
REPAIR-Bench provides both the HRI and Medical HRI communities with a standardized framework for (1) evaluating robot failures and (2) building transparent, adaptive, and trustworthy recovery systems.

\end{abstract}

% Another version for consideration: Understanding how users perceive and respond to failures is essential for robust robot systems. Prior work, however, largely studies individual failures in isolation, offers limited modeling of their temporal dynamics, and emphasizes binary failure detection with little insight into users’ preferred failure recovery. We present REPAIR-Bench, a benchmark for embodied failure understanding that integrates trial-level failure progression, multimodal behavioral signals, and user preferences for recovery. Built on 214 interaction trials from 41 participants, the benchmark spans four induced failure types and provides synchronized facial action units, head pose, speech transcripts, and post-interaction affect and recovery reports.
% REPAIR-Bench defines three tasks: sliding-window failure classification, temporal boundary detection, and recovery strategy prediction. We also evaluated an HRNN and an SLM on these tasks and found [placeholder].
% \end{abstract}


\section{Introduction}

Robots are increasingly developed for high-stakes environments, such as emergency departments~\cite{tanjim2025human}. However, even well-engineered systems are prone to failures in real-world settings, due to perception errors, time pressure, misinterpretation of user intent, or environmental uncertainty~\cite{honig2018understanding}.

Characterizing failures and user responses is critical, as effective handling influences task completion, trust, and long-term acceptance of robots in the workplace. ~\cite{honig2018understanding}.%Characterizing how failures unfold and how users perceive and respond to them is of the utmost importance, as effective failure handling influences not only task completion, but also users' trust, willingness to re-engage, and long-term acceptance of robots~\cite{honig2018understanding}.

\begin{figure}[t]
\centering
\includegraphics[width=1.0\linewidth]{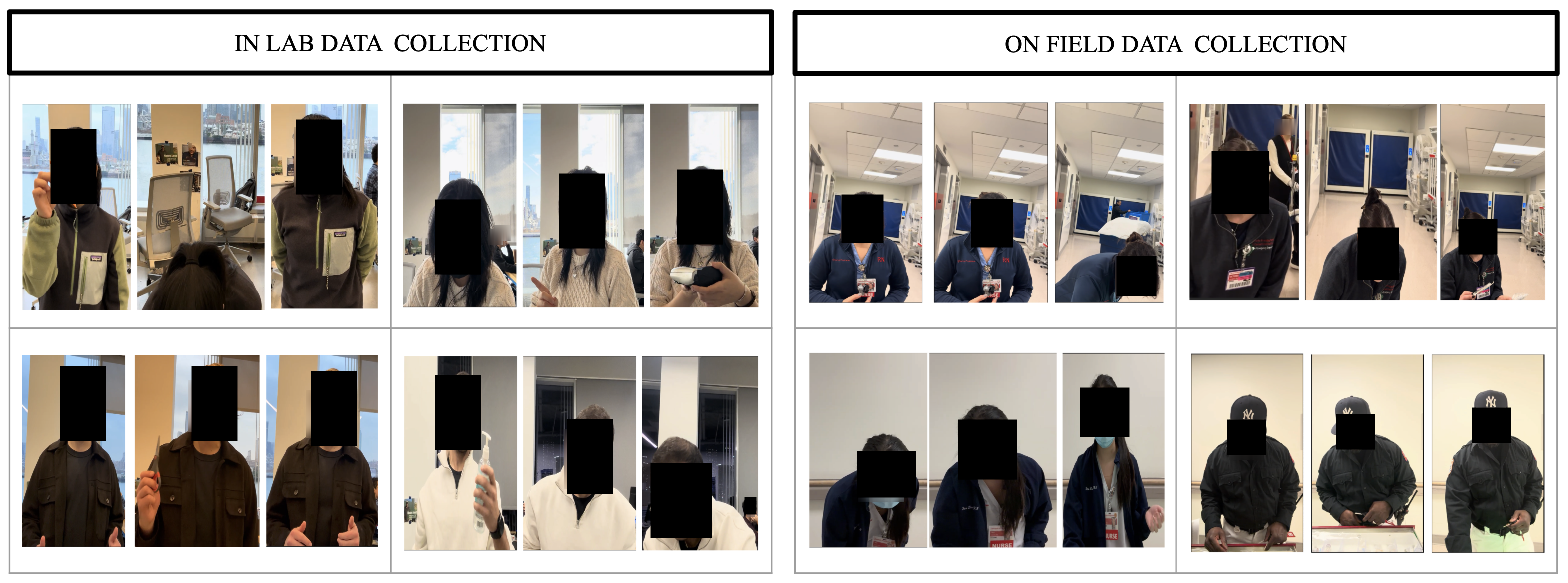}
\caption{RFM-HRI data collection environment at an academic laboratory setting (left) and at the hospital (right) \cite{rfm-hri}. }
\label{fig:setup}
\end{figure}

HRI research increasingly focuses on recognizing task failures through user behavior. System breakdowns disrupt interaction in measurable ways: the ERR@HRI dataset~\cite{cao2025err} highlights delays and robot-initiated interruptions as primary causes of rupture, while Kontogiorgos et al.~\cite{kontogiorgos2021systematic} identify multimodal markers like gaze shifts and acoustic changes. Nonverbal cues often precede explicit verbal complaints, serving as early failure indicators~\cite{bremers_using_2024}.

Structured multimodal analysis has been advanced by the Response-to-Errors dataset, which uses OpenFace-extracted action units (AUs)~\cite{stiber2023social}, and ERR@HRI 2.0, which includes corrective intent in conversational settings. Recent work leverages representation learning~\cite{prisca2024errhri}, time-series pipelines~\cite{pipeline2024}, and alignment strategies~\cite{alignment2024} to detect errors from user reactions alone. Further extensions have moved beyond technical failures to detect social breakdowns and repair attempts~\cite{beyondtech2024}, establishing a strong foundation for conversational failure detection.

Despite these advances, several gaps remain.
First, most benchmarks treat failures as independent samples rather than modeling cross-trial history or accumulation effects~\cite{khanna2025reflex}.
Second, existing datasets primarily support binary detection rather than multi-class failure categorization.
Third, while prior work has examined recovery strategies, these are typically researcher-defined rather than predicted from user behavior.
Finally, there is a lack of integrated benchmark frameworks connecting detection, understanding, and recovery decision-making within a single evaluation setting.

\begin{figure}[t]
\centering
\includegraphics[width=1.0\linewidth]{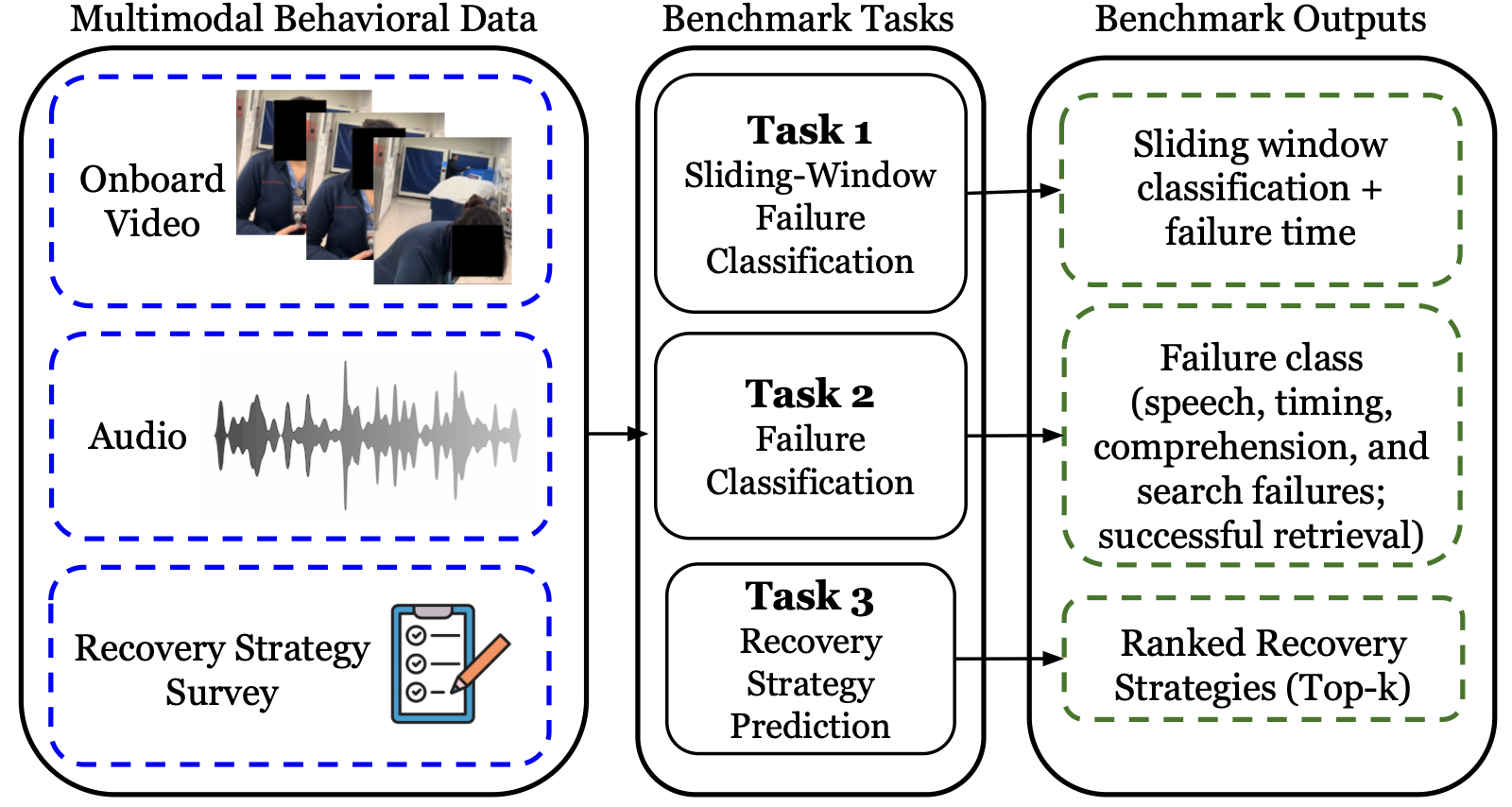}
\caption{REPAIR-Bench is multimodal framework for temporal robot failure detection, failure classification, and recovery strategy prediction.}
\label{fig:overview}
\end{figure}

We introduce the Benchmark for Robot Error Perception And Interaction Recovery (REPAIR-Bench), the first HRI benchmark to jointly address failure classification, localization, and recovery prediction (Figure \ref{fig:overview}). Using the RFM-HRI dataset (Figure \ref{fig:setup}), we evaluate a Hierarchical Recurrent Neural Network (HRNN) for temporal detection and fine-tuned Small Language Models (SLMs) for recovery. Our results show that hierarchical modeling of cross-session history improves performance and that QLoRA-based instruction tuning is essential for structured output tasks.

\textbf{Contributions. }
REPAIR-Bench captures failures that occur while users interact with a physical robot to complete a physical retrieval task using a novel robot platform.
We introduced three novel tasks, each designed to address key gaps mentioned above. 
The first task focuses on failure time detection and captures the accumulation of failure signals across multiple sessions. Importantly, it relies exclusively on visual behavioral signals, since they often provide earlier indicators of failure than verbal responses of users. In addition, audio inputs that include the robot’s speech would allow models to rely on the moment the robot begins speaking to infer when the failure occurred, while our focus is on capturing when and how external behavioral cues signal the emergence of a failure. 
The second task focuses on distinguishing between failure categories using solely visual cues of a user's reaction, and provides critical contextual information about why the interaction broke down, which is essential for designing appropriate system responses. 
The third task predicts user-preferred recovery strategies using interaction transcripts and the failure context. The benchmark enables research on adaptive recovery behaviors that align with human expectations rather than designer-defined rules.
Finally, REPAIR-Bench enables systematic investigation of how robots can detect failures, understand their underlying context, and select recovery strategies that help restore user confidence and trust after interaction breakdowns.

\begin{figure*}[t]
\label{fig:robot}
\centering
\includegraphics[width=1.0\textwidth]{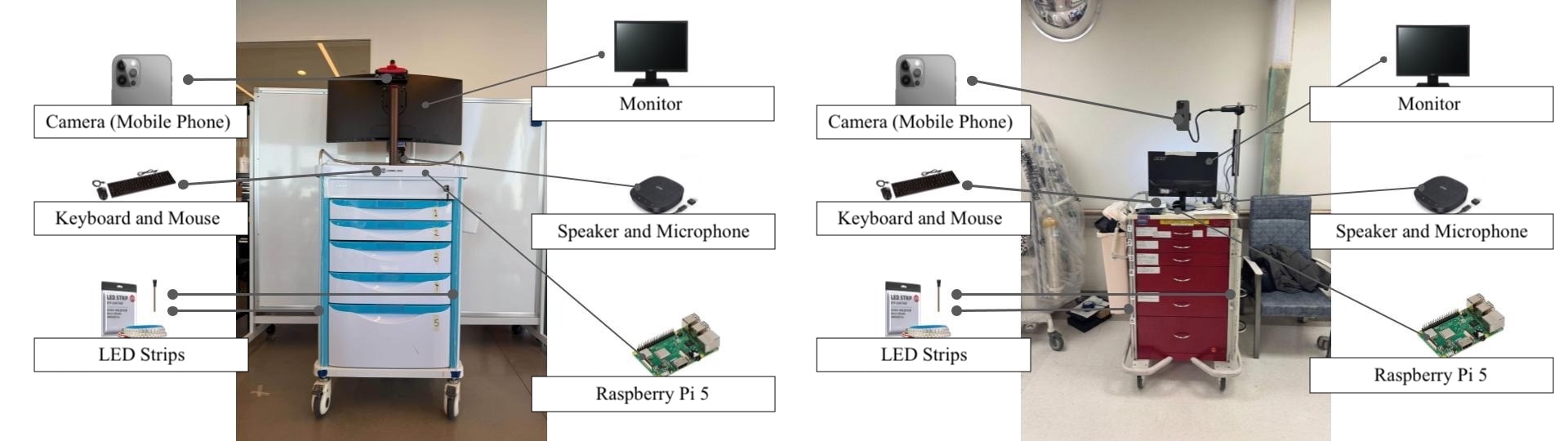}
\vspace{0.5em}
\begin{minipage}{0.48\textwidth}
\centering
\small
\textbf{In Lab Data Collection}
\end{minipage}
\hfill
\begin{minipage}{0.48\textwidth}
\centering
\small
\textbf{In Field Data Collection}
\end{minipage}

\caption{The RFM-HRI dataset \cite{rfm-hri} captures people performing object retrieval tasks with a medical robotic crash cart. CrashCart is a hospital-deployed mobile manipulator designed to support rapid crash-cart retrieval and handoff. It provides multimodal guidance via both visual and audio cues (e.g., light-emitting diode (LED) plus spoken instructions) to help users quickly understand the robot’s state and next actions during time-critical workflows.}
\label{fig:crashcart}
\end{figure*}

% \section{Methodology}

\section{Dataset and Preprocessing}

REPAIR-Bench builds on RFM-HRI, a multimodal dataset of HRIs during crash-cart item retrieval~\cite{rfm-hri}, one of the only publicly available datasets that captures both robot failures and preferred recovery behaviors within a medical context.

The dataset records interactions with a medical crash cart, a mobile cart used in hospital emergency rooms to store emergency medications and life-saving equipment for rapid access during critical events. The RFM-HRI dataset introduces a semi-autonomous crash-cart robot that guides users through retrieving specific items. The robot is a retrofitted medical crash cart equipped with LED indicators that highlight drawer locations, a text-to-speech interface for verbal guidance, and an egocentric camera that captures the interaction (Figure~\ref{fig:crashcart}).

The dataset contains 214 interaction samples from 41 participants collected in both laboratory and hospital settings. The original study included recordings from 41 participants, each completing five trials comprising four experimentally induced robot failure conditions in random order (i.e., speech failure, timing failure, comprehension failure, and search failure) and one error-free success condition (Figure~\ref{fig:setup}). For each trial, the dataset includes synchronized video, audio, and task transcripts. Following each interaction, participants completed a post-interaction survey in which they reported (1) their emotional reactions to the trial and (2) their preferred recovery strategy (e.g., apology and clarification, step-by-step guidance, escalation to a human teammate).

Each sample in the processed dataset includes: (i) facial Action Units (AUs) (18 intensity features), gaze estimates, and 3D head pose (pitch, yaw, roll) extracted via Google MediaPipe; (ii) 3D facial landmarks (468 keypoints per frame); (iii) time-stamped speech transcripts produced using the Whisper Automatic Speech Recognition (ASR) model; and (iv) post-trial survey responses capturing dominant emotion, Self-Assessment Manikin (SAM) affective ratings, and recovery strategy preferences \cite{bradley1994measuring}.

Building on this foundation, we introduce extensions that transform the dataset into a unified benchmark for fine-grained failure modeling and evaluation. First, we standardized four failure types (i.e., speech, timing, comprehension, search, and success) and annotate each trial accordingly (see Section \ref{subsec:data_label}). For trials involving failures, we provided precise onset and offset timestamps, enabling temporal localization tasks. Second, we extracted multimodal behavioral features using MediaPipe-based pose and gaze tracking, along with complementary linguistic features, to quantify users' moment-by-moment reactions. Third, we preprocessed transcripts to support recovery modeling.

\subsection{Sample Exclusions.} We excluded 30 samples that did not meet benchmarking criteria. Specifically, some trials contained duplicated or near-identical transcripts across samples, which would artificially inflate model performance under transcript-based recovery prediction. Additional trials were excluded when transcript content did not provide sufficient linguistic evidence to localize or characterize the failure event, preventing reliable supervision for recovery modeling. After exclusion, the final analysis set contains of 184 samples. %A list of excluded sample IDs will be provided in the released benchmark documentation.

\subsection{Temporal Failure Annotations.} To evaluate temporal failure localization, we annotate failure trials with an onset and offset timeframe. The frame is computed as $f_{\text{onset}} = \text{FPS} \times (t_{\text{fail}} - t_{\text{start}})$
%\begin{equation}
%f_{\text{onset}} = \text{fps} \times (t_{\text{fail}} - t_{\text{start}})
%\label{eq:onset}
%\end{equation}
where frames per second (FPS) is the ratio of frame count to trial duration from each sample's features, $t_{\text{fail}}$ is the failure start times from system logs, and $t_{\text{start}}$ is the trial start times. %each sample's MediaPipe features, $t_{\text{fail}}$ is the failure start timestamp from system logs, and $t_{\text{start}}$ is the trial start timestamp. 
These annotations define the ground truth for detection latency evaluation.

\subsection{MediaPipe Feature Preprocessing.} Frames with failed face detection are excluded. Missing values in retained frames are filled via linear interpolation. Following Fast Fourier Transform (FFT) analysis we applied bidirecional second order Butterworth filter (6 Hz) to mediapipe time series to reduce high-frequency noise while preserving behaviorally relevant dynamics. Features are normalized by range per feature per sample. Features identified as unreliable based on near-zero variance across conditions are excluded. The resulting input has 21 channels per frame (18 AU + 3 head pose).

\subsection{Transcript Preprocessing.} 
Transcripts are used solely for the recovery modeling component. We remove timestamps, retaining only the natural-language dialogue, since temporal boundary information is unnecessary for recovery prediction. No paraphrasing or restructuring is performed; the original conversational content is preserved.
We then construct a supervised instruction-tuning dataset with explicit input–output pairs. Each input consists of a structured prompt defining the robot’s role, the transcript, the success/failure type label, and a fixed list of candidate recovery strategies. The output is a comma-separated list of the gold-standard recovery strategies based on user-reported preferences (or an empty response if the interaction succeeded). This formulation casts recovery prediction as a constrained generative ranking problem with standardized output formatting.

% \subsection{Data Annotation and Reliability.} To ensure the quality and consistency of the labels, annotations were produced by four trained members of the research team following a shared, written annotation protocol. The dataset was independently coded by these annotators, who reviewed the interaction transcripts alongside their available timestamps. They were tasked with categorizing the specific failure types and identifying the exact beginning and end of failure events based on the conversational flow. Consistency was supported through the use of a common definition for failure onset, established as the exact timestamp where explicit, predefined robot utterances appeared in the dialogue. These explicit signals corresponded directly to the four failure conditions: ``Open the drawer'' (Speech Failure), ``Sorry, I am delayed...'' (Timing Failure), ``I think the item is in drawer \_\_\_'' (Search Failure), and ``I did not understand that request.'' (Comprehension Failure). Any discrepancies among the annotators were resolved via consensus discussion to establish the final ground-truth labels used for model evaluation.

\subsection{Data Annotation and Reliability.} 
\label{subsec:data_label}

To ensure the quality and consistency of the labels, annotations were produced by four trained research team members following a shared written protocol. Annotators independently coded the interaction transcripts with timestamps, categorizing failure types and marking exact onset and offset of events. Temporal consistency was supported by defining failure onset as the timestamp of explicit robot utterances corresponding to the four failure conditions: ``Open the drawer'' (Speech), ``Sorry, I am delayed...'' (Timing), ``I think the item is in drawer \_\_\_'' (Search), and ``I did not understand that request'' (Comprehension). Discrepancies among annotators were resolved by 100\% consensus.

\section{Baseline Models and Evaluation}

\subsection{Task 1. Failure Detection}

\begin{figure}[t]
\centering
\includegraphics[width=1.0\linewidth]{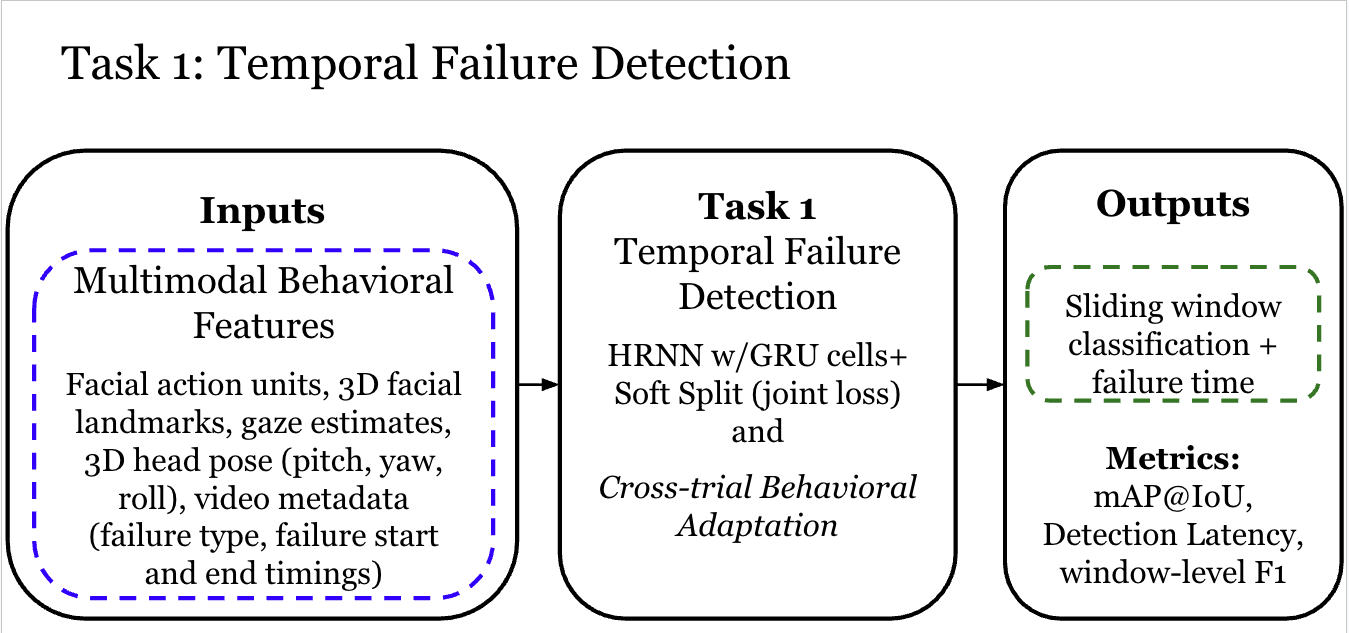}
\caption{Task 1: Temporal failure detection using 1) Hierarchical Recurrent Neural Networks (HRNN) with Bidirectional Long Short Term Memory (BiLSTM) and 2) HRNN with Gated Recurrent Unit (GRU) cells using multimodal visual features as input and the failure transition moment as output.}
\label{fig:task1}
\end{figure}

\textbf{Task Definition.}
Given sequential visual features extracted at the frame level (MediaPipe facial AUs, facial landmarks, head pose, and gaze), the objective is to detect the temporal moment at which a robot can infer that its response has failed based on the user’s behavioral reaction within an interaction session while accounting for cross-trial dependencies (see Figure \ref{fig:task1}).

Inputs: Each trial is represented as a temporal sequence of sliding windows computed over frame-level MediaPipe features (facial landmarks, AUs, head pose, and gaze, selectively used). Within each window, summary statistics (mean, min, max, standard deviation) of these signals are computed, producing a fixed-dimensional feature vector per window that characterizes the user's visual behavior over that interval.

Outputs:
The model produces two forms of prediction per trial: (i) a binary label for each window indicating whether the user's behavior corresponds to a pre-failure state (before the transition) or a post-failure state (after the transition); (ii)
transition moment estimation: a single temporal prediction indicating when the user's behavioral shift occurs, expressed in seconds relative to trial onset.

Ground Truth:
We define the failure offset as the ground-truth transition moment. This choice reflects the nature of the interaction: the robot's response consists of a sequence of words, and the first word or the beginning of the response is not itself an error by definition, but the failure lies in the content of the full response. Only once the robot has finished speaking can the response be understood and defined as incorrect (failed). The failure offset (end of robot's failed response) therefore marks the earliest point at which a behavioral shift can reasonably be expected. For success trials, no transition exists, and the entire trial is labeled as pre-failure. Window-level ground-truth labels are derived accordingly: any window whose start time falls at or after the failure offset is labeled as post-failure; all preceding windows are labeled as pre-failure (because each session captures a user's request, robot's successful or failed response, and the user's reaction—within a short time span, after which no additional interaction events occur within the same trial.)

\textbf{Architecture.}
Our best-performing model is a two-level HRNN with GRU cells, operating at window and session-level temporal scales, and trained with an auxiliary localization objective.

At the window level, a single-layer GRU receives the full sequence of window feature vectors for one trial, producing a hidden state at each timestep. Each hidden state is passed through layer normalization, dropout, and a linear classifier to produce per-window pre/post-failure logits. The GRU’s final hidden state serves as a trial embedding.

A GRU cell maintains a history vector that is updated after each completed trial by ingesting the trial embedding concatenated with a binary failure indicator. 
Before each new trial, the history vector is projected through a learned bridge layer (linear + tanh) and used to initialize the window-level GRU's hidden state $h_0$, conditioning the window-level processing on the participant's prior interaction history.

For localization, the per-window logits are passed through a softmax to obtain continuous pre-failure probabilities, enabling end-to-end backpropagation. The predicted transition index is obtained by summing these probabilities across windows and is supervised with a Huber regression loss against the ground-truth index, which is robust to outlier trials with large localization errors. The predicted index is converted to seconds via the corresponding onset time.

We explored multiple alternative architectures and temporal  localization approaches. A stateful GRU variant carried its hidden state across trials via a learned bridge layer, treating cross-trial context as implicit memory within a single recurrent module rather than as an explicit hierarchical level. A three-level hierarchical model separated the trial encoder and window classifier into distinct GRU modules, adding a dedicated encoder for trial embeddings. We also evaluated flat baselines (GRU and BiLSTM) with hand-crafted cross-trial features (mean statistics, failure counts, and post-failure ratios from preceding trials) concatenated to the input; while competitive, these rely on ground-truth labels from prior trials at inference. A Random Forest (RF) on window-level features served as a non-sequential reference. %Details are reported below.

\textbf{Training Procedure.}
Prior to training, 30 samples with inaccurate timestamps or conflicting multi-label annotations were excluded, yielding 184 usable trials across 38 participants. Within each trial, Not a Number (NaN) values were interpolated, each channel was smoothed with a fourth-order Butterworth low-pass filter (5\,Hz cutoff), and per-trial min–max normalization to [0,1]. %IV-order Butterworth low-pass filter (5\,Hz cutoff), and per-trial min–max normalization to [0,1] was applied. 
After windowing, features were globally z-score standardized with residual NaNs handled by mean imputation. Feature selection, guided by distributional analysis and preliminary Random Forest permutation importance, led to excluding depth (z-) coordinates from pose and landmark modalities and dropping the full 468-point facial landmarks entirely. Within each window, the the final feature vector includes the mean, min, max, standard deviation per channel. %four summary statistics (mean, min, max, standard deviation) per channel form the final feature vector. 

Beyond the architectural variants described above, we conducted an extensive experimental search over training configurations, including loss formulations (soft-split Mean Square Error (MSE) with separation penalty, summation-based Huber, and cross-entropy alone), loss weighting schedules ($\lambda_{\mathrm{loc}} \in [0.5, 3.0], \delta \in [2.0, 6.0]$), temporal granularities (2- and 5-frame windows stride vs.\ long 3\,s windows with 0.5\,s stride), dropout rates, hidden dimensions, learning rates, and optimizer choices. Below we report the training details of the best-performing configuration: the two-level HRNN with 3\,s windows and summation-based Huber localization.

Data are split at the participant level (~75\%/25\%), ensuring all trials from a given participant appear in a single partition. The HRNN is trained with participant-ordered iteration: within each epoch, participants are shuffled, but trials within each participant are processed strictly in temporal order so that the session-level hidden state accumulates a coherent interaction history. The session-level state is detached between participants but flows uninterrupted across trials.% within a participant.

The training loss combines window-level cross-entropy with inverse-frequency class weighting and label smoothing (0.05), and a Huber localization term ($\lambda_{\mathrm{loc}}=2.0, \delta=4.0$). We optimize with AdamW ($\eta{=}2{\times}10^{-4}$, weight decay $10^{-4}$) and cosine annealing to $10^{-6}$, clipping gradient norms at 1.0. Training runs up to 150 epochs with early stopping (patience 30) on held-out participants; model selection maximizes a composite score of detection rate, temporal accuracy, and success-trial false-positive suppression.

Several challenges are handled. First, the dataset contains inconsistencies (e.g., multi-label samples), which reduces the effective training size. Second, the magnitude of noise in facial expression descriptors is often comparable to the subtle facial changes associated with user reactions. Third, the visual feature space is high-dimensional (1467 descriptors), while the most informative modality—speech—was intentionally excluded in task definition. As a result, the task relies solely on visual signals, where even human observers often struggle to detect clear facial differences following failure events. We address these issues through careful preprocessing, feature selection, model architecture and tuning, as described above. Temporal smoothing and short-term dynamics are captured using sliding windows with window-level statistical features rather than per-frame inputs, which reduces frame-level noise.

\textbf{Evaluation Metrics.}
We evaluate performance at two levels. At the window level, we report precision, recall, and F1 (macro-averaged) for the binary classification \cite{vanrijsbergen1979}. %Because the exact boundary between pre- and post-failure behavior is inherently ambiguous (a participant's visible reaction may lag the failure event by several seconds) we additionally report tolerant F1, which credits a window prediction as correct if the ground-truth label appears within a $\pm \tau$ s temporal neighborhood of the same trial, reported at  $\tau \in {0, 0.5, 1, 2, 3}$\,s. 

At the trial (sample) level, mean and median absolute localization error (in seconds) characterize the accuracy for each failure type. We also report F1 at temporal Intersection-over-Union (tIoU) thresholds of 0.1, 0.25, 0.5, and 0.75, comparing the predicted and ground-truth post-failure segments. All metrics are computed on held-out participants.

\subsection{Task 2: Failure Classification}

\textbf{Task Definition.} 
The task is condition-type classification (comprehension, speech, timing, search failures, and success) of the sample using visual time-series features derived via MediaPipe (facial AUs, 3D facial landmarks, gaze estimates, 3D head pose (pitch, yaw, roll)), taking into account user's previous session experience.

\begin{figure}[h]
\centering
\includegraphics[width=1.0\linewidth]{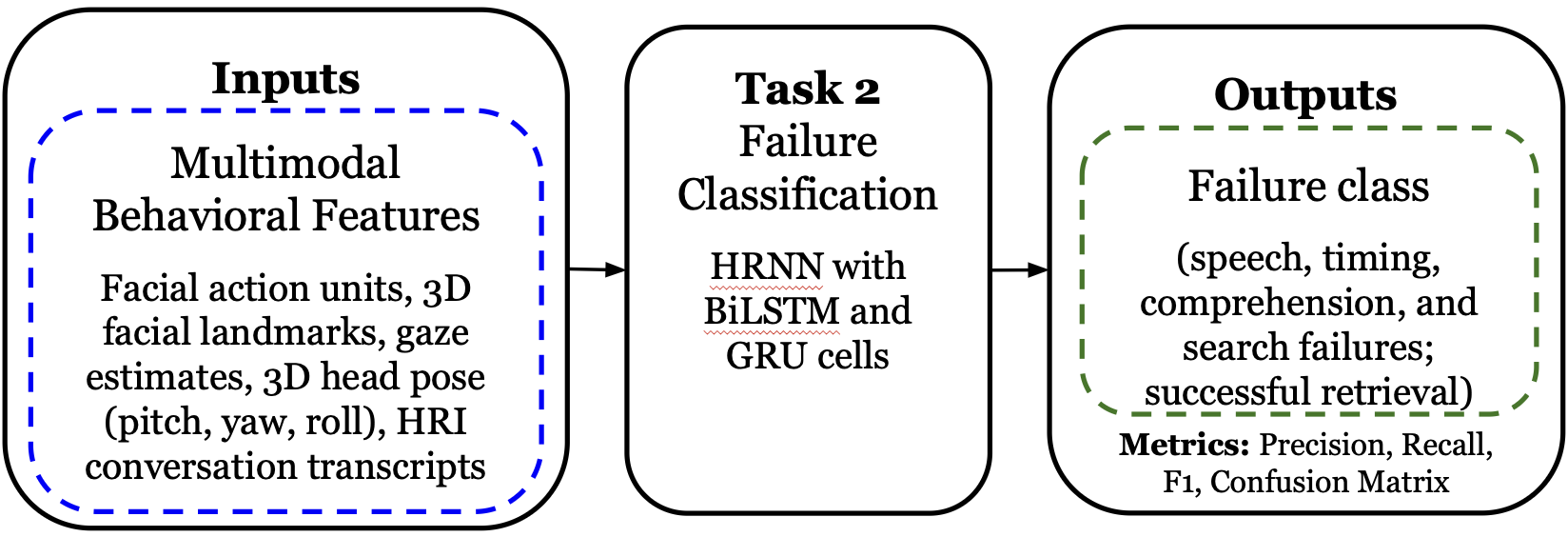}
\caption{Task 2: Failure classification using Hierarchical Recurrent Neural Networks (HRNN) with Bidirectional Long Short Term Memory (BiLSTM) and Gated Recurrent Unit (GRU) cells using multimodal features as input and failure classes as output. }
\label{fig:task2}
\end{figure}

\textbf{Architecture.} We employed a two-level, session-aware temporal model (see Figure \ref{fig:task2}). 
At the cross-session level, a bidirectional LSTM encodes frame-wise features within each interaction. 
A failure-aware temporal attention mechanism aggregates these frame representations into a single session embedding. 
At the current session level, a unidirectional GRU processes the sequence of trial embeddings in chronological order to capture dependencies across interactions. 
The final prediction is obtained by combining the trial embedding with its session-level context and passing it through a multi-layer perceptron.

 \textbf{Training Procedure.}
Data are split at the participant level into training and held-out participant sets (approximately 75\%/25\%), ensuring that all trials from a participant belong to a single split to prevent information leakage across sessions. Feature normalization parameters are estimated using only the training participants and then applied to the held-out participants. Within the training set, trials belonging to the same participant are processed sequentially in their original temporal order to preserve interaction context across trials. Each trial is represented as a variable-length sequence of frame-level visual features (padded or truncated to a fixed maximum length during batching), and the model predicts a single failure-type label per trial.

\textbf{Evaluation Metrics.}
We evaluate failure-type classification using standard multi-class metrics: precision, recall, and F1-score similar to prior work \cite{vanrijsbergen1979}. 

\subsection{Task 3: Recovery Strategy Prediction}

\begin{figure}[h]
\centering
\includegraphics[width=1.0\linewidth]{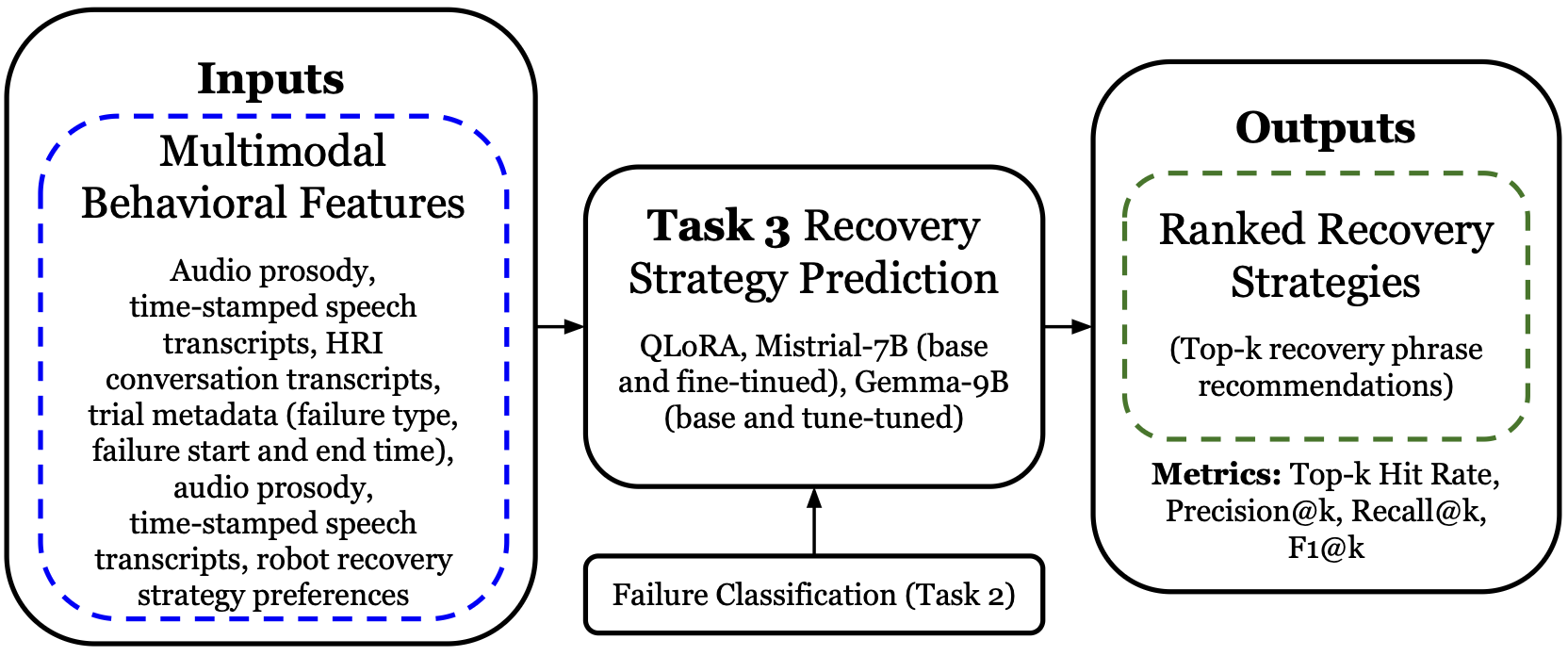}
\caption{Task 3: Recovery strategy prediction using QLoRA \cite{qlora}, Mistrial-7B \cite{jiang2023mistral7b}, Gemma-9B \cite{GoogleGemma2024} with multimodal features, robot recovery strategy preferences, base and fine-tuned Small Language Models (SLMs) as inputs and ranked recovery strategies as output.}
\label{fig:task3}
\end{figure}

\textbf{Task Definition.} 
The third model addresses recovery recommendation, following a failure event (see Figure \ref{fig:task3}). Given (i) the transcript of a robot–user interaction, (ii) a failure label (failure type or success), and (iii) a predefined set of candidate recovery strategies, the model must generate the top-$k$ most appropriate recovery actions when a failure occurs. If the interaction is labeled as successful, the model is required to return an empty response.

We frame this as a \textit{constrained generative ranking problem}: outputs must consist strictly of comma-separated strategy names from the candidate vocabulary, without explanations or additional text. The first $k$ generated strategies are treated as an ordered prediction set.

\textbf{Architecture.} We evaluate both base instruction models and domain-adapted variants fine-tuned using Quantized Low Rank Adaptation (QLoRA) \cite{qlora}: (i) Mistral-7B (base), (ii) Mistral-7B fine-tuned, (iii) Gemma-9B (base), and (iv) Gemma-9B fine-tuned \cite{jiang2023mistral7b, GoogleGemma2024}. 

Fine-tuned models are adapted using QLoRA on an NVIDIA A100 GPU (40GB) with 4-bit normal float (NF4) quantization \cite{qlora}, double quantization, and bfloat16 compute dtype for memory-efficient training. Low-Rank Adapters (LoRA) are inserted into attention ($q$, $k$, $v$, $o$) and Multi Layer Perceptron (MLP) layers, allowing domain specialization without updating full model weights.

\textbf{Training Procedure.} The dataset is randomly split into train (60\%), validation (20\%), and test (20\%) subsets. Prompts are tokenized with truncation and padded to a maximum sequence length of 810 tokens. Labels are set equal to input IDs to optimize a causal language modeling objective over the full prompt–completion sequence. Training uses paged AdamW (8-bit) with a learning rate of $5\times 10^{-5}$, per-device batch size of 2, and a maximum of 250 steps, or around 4.5 epochs. Gradient checkpointing is enabled. Only LoRA adapter parameters are updated.

Training prompts are defined as the 1) robot’s role (medical crash cart assistant (see \ref{fig:crashcart})), 2) output formatting constraints, 3) interaction transcript, 4) failure label, and 5) candidate recovery strategy. This structured instruction-tuning ensures consistent supervision and reduces output variance. During evaluation, recovery generation is deterministic, and the first $k$ comma-separated strategies are extracted as a predicted set for Top-$k$ evaluation.%(greedy decoding, beam size = 1), and the first $k$ comma-separated strategies are extracted as a predicted set for Top-$k$ evaluation.

\textit{Early stopping:} Mistral-7B with QLoRA was early-stopped after 110 steps (2 epochs), and Gemma-9B with QLoRA after 220 steps (4 epochs), as training losses had converged, ensuring efficient adaptation without overfitting. Both models achieve stable Top-$k$ performance without signs of overfitting, indicating efficient adaptation through QLoRA with 4-bit NF4 quantization and LoRA adapters.

\textbf{Evaluation Metrics.} Recovery prediction produces one or more results. Thus, we assessed recovery prediction using set-based Top-$k$ metrics for $k \in \{1,2,3,4,5\}$:
(i) Hit@$k$: at least one gold-standard strategy appears in the top-$k$ predictions; \cite{vanrijsbergen1979}
(ii) Precision@$k$: fraction of predicted strategies in the top-$k$ that are correct; \cite{vanrijsbergen1979}
(iii) Recall@$k$: fraction of gold strategies recovered in the top-$k$;  \cite{vanrijsbergen1979} 
(iv) F1@$k$: harmonic mean of Precision@$k$ and Recall@$k$. \cite{vanrijsbergen1979}
This evaluation captures partial correctness and rank sensitivity in settings where multiple recovery strategies may apply.

\section{Results} 

We present results for our two models across the three benchmark tasks spanning the failure lifecycle: sliding-window failure classification (Task 1), temporal failure detection (Task 2), and recovery preference prediction (Task 3).  Tasks 1 and 2 measure a model’s ability to detect and classify failures, while Task 3 evaluates the downstream impact of these predictions through recovery strategy selection. For clarity, results are grouped by task, with both model comparisons and evaluation metrics highlighted.

\subsection{Task 1: Failure Detection} 

Table~\ref{tab:window-classification} report window-level classification and tolerant F1 across selected best-performance architectures. The two-level HRNN with 3s windows achieves the best performance, with macro F1 of 0.80 (strict) rising to 0.87 at $\pm$3s tolerance, and AUC of 0.87, outperforming by 0.15 standard GRU baseline. Table~\ref{tab:localization} reports transition moment (failure offset) localization on held-out participants: the model detects all 30 failure trials with a median absolute error of 2.97 s and mean mIoU of 0.781. The overall (mean) signed error of $-$0.51s indicates a slight early-prediction tendency.

% Window-Level Classification (Pre- vs. Post-Failure)
%%%%%%%%%%%%%%%%%%%%%%%%%%%%%%%%%%%%%%%%%%%%%%%
\begin{table}[t]
\centering
\caption{Window-level binary classification (pre- vs.\ post-failure) $^{\dagger\dagger}$ ($\uparrow$ means higher is better).}
\label{tab:window-classification}
\setlength{\tabcolsep}{3.5pt}
\footnotesize
\begin{tabular}{@{}llcccccc@{}}
\toprule
Model & Win & Acc $\uparrow$  & BAcc $\uparrow$ & Prec $\uparrow$ & Rec $\uparrow$ & F1 $\uparrow$ & AUC $\uparrow$ \\
\midrule
\multicolumn{8}{@{}l}{\textit{Flat baselines}} \\[2pt]
RF                              & 3\,s & 0.61 & 0.62 & 0.56& 0.70 & 0.61 & 0.66 \\
RF w/hist$^\dagger$          & 3\,s & 0.64 & 0.64 & 0.59 & 0.65 & 0.63 & 0.69 \\
GRU                             & 3\,s & 0.68 & 0.68 & 0.67 & 0.70 & 0.68 & 0.72 \\
GRU w/hist.$^\dagger$         & 3\,s & 0.69 & 0.69 & 0.70 & 0.63 & 0.69 & 0.72 \\
BiLSTM w/hist.$^\dagger$      & 3\,s & 0.69 & 0.69 & 0.70 & 0.67 & 0.69 & 0.71 \\
\midrule
\multicolumn{8}{@{}l}{\textit{Hierarchical models}} \\[2pt]
HRNN (3 level)                & 3\,s & 0.66 & 0.66 & 0.64 & 0.73 & 0.66 & 0.69 \\
HRNN (2 level)                  & 3\,s & \textbf{0.80} & \textbf{0.80} & \textbf{0.74} & \textbf{0.85} & \textbf{0.80} & \textbf{0.87} \\
HRNN (2 level)                           & 5\,f & 0.70 & 0.71 & 0.66 & 0.78 & 0.70 & 0.77 \\
\bottomrule
\end{tabular}

\vspace{4pt}
{\footnotesize
Win: window size (3\,s = 3-second, 5\,f = 5-frame).

$^\dagger$\, \emph{History-aware models using cross-trial features concatenated to input;
requires ground-truth labels from prior trials at inference.}

$^{\dagger\dagger}$ \emph{Note: $\uparrow$ indicates a higher metric value is better and $\downarrow$ indicates a lower value is better.}
}
\end{table}

% % Table 2: Tolerant F1 

% \begin{table}[t]
% \centering
% \caption{Tolerant F1 for window classification under varying temporal
% tolerances. A prediction is counted as correct if the ground-truth
% label appears within $\pm\tau$ of the predicted window $^{\dagger\dagger}$ ($\uparrow$ means higher is better). }
% \label{tab:tolerant-f1}
% \setlength{\tabcolsep}{4pt}
% \footnotesize
% \begin{tabular}{@{}llcccc@{}}
% \toprule
% Model & Win & Strict F1 $\uparrow$ & $\pm$1\,s F1 $\uparrow$ & $\pm$2\,s F1 $\uparrow$ & $\pm$3\,s F1 ss$\uparrow$ \\
% \midrule
% \multicolumn{6}{@{}l}{\textit{Flat baselines}} \\[2pt]
% RF                              & 3\,s & 0.61 & 0.64 & 0.66 & 0.67 \\
% RF w/hist.$^\dagger$          & 3\,s & 0.63 & 0.66 & 0.68 & 0.70 \\
% GRU                             & 3\,s & 0.68 & 0.70 & 0.73 & 0.75 \\
% GRU w/hist.$^\dagger$         & 3\,s & 0.69 & 0.71 & 0.74 & 0.76 \\
% BiLSTM w/hist.$^\dagger$      & 3\,s & 0.69 & 0.72 & 0.74 & 0.76 \\
% \midrule
% \multicolumn{6}{@{}l}{\textit{Hierarchical models}} \\[2pt]
% HRNN (3 level)               & 3\,s & 0.66 & 0.69 & 0.71 & 0.73 \\
% HRNN  (2 level)                 & 3\,s & \textbf{0.80} & \textbf{0.82} & \textbf{0.85} & \textbf{0.87} \\
% HRNN (2 level)                            & 5\,f & 0.70 & 0.71 & 0.71 & 0.72 \\
% \bottomrule
% \end{tabular}

% \vspace{4pt}
% {\footnotesize Win: window size (3\,s = 3-second, 5\,f = 5-frame).

% $^\dagger, ^{\dagger\dagger}$\,See Table~\ref{tab:window-classification} footnote. }

% \end{table}

% Transition Moment Localization (HRNN)

\begin{table}[t]
\centering
\caption{Transition moment localization for the HRNN (2 level, 3\,s windows)
detection rate. MAE/MedAE: mean and median absolute
error in seconds. Mean\,E: signed mean error (negative = early prediction).
mIoU: mean temporal intersection-over-union with the ground-truth
post-failure segment ($\uparrow$ means higher is better).}
\label{tab:localization}
\setlength{\tabcolsep}{4pt}
\footnotesize
\begin{tabular}{@{}lccccc@{}}
\toprule
Failure type & $N$ & Det\% $\uparrow$ & MedAE\,(s) $\downarrow$ & Mean\,E\,(s) $\downarrow$ & mIoU $\uparrow$ \\
\midrule
Comprehension & 7 & 100 & 2.06 & --- & 0.89 \\
Search        & 8 & 100 & 4.25 & --- & 0.82 \\
Speech        & 7 & 100 & 3.23 & --- & 0.74 \\
Timing        & 8 & 100 & 4.38 & --- & 0.68 \\
\midrule
\textbf{Overall} & \textbf{30} & \textbf{100} & \textbf{2.97} & \textbf{$-$0.51} & \textbf{0.78} \\
\bottomrule
\end{tabular}

\vspace{4pt}
{$^{\dagger\dagger}$\,See Table~\ref{tab:window-classification} footnote. }

\end{table}

\subsection{Task 2: Failure Classification}
We reported the best performance models among experiments. Table~\ref{tab:overall} shows that HRNN achieves the best overall performance (macro F1 = 0.44), improving over BiLSTM-GRU (F1 = 0.38) despite having the same number of parameters. Table~\ref{tab:perclass} indicates that gains are driven mainly by much stronger performance on the Success class (F1 = 0.93) and improved detection of Speech failures, while Comprehension and Timing failures remain harder to distinguish. Overall, the results suggest that failure type classification is feasible but still challenging across several classes. 
This is consistent with the statistical analysis results showing a modest association between self-reported emotion and specific failure condition: a chi-square test of independence across the five conditions: $\chi^2(48) = 112.76, p < .001$, with an effect size of Cramér's $V = 0.386$.

\begin{table}[t]
\centering
\caption{Overall macro-averaged classification performance.}
\label{tab:overall}
\small
\begin{tabular}{@{}lcccc@{}}
\toprule
\textbf{Model} & \textbf{Params} & \textbf{P $\uparrow$} & \textbf{R $\uparrow$} & \textbf{F1 $\uparrow$} \\
\midrule
BiLSTM-GRU & 1{,}012K & 0.42 & 0.41 & 0.38 \\
HRNN            & 1{,}012K & \textbf{0.50} & \textbf{0.44} & \textbf{0.44} \\
\bottomrule
\end{tabular}

\vspace{2pt}
\end{table}

\begin{table}[t]
\centering
\caption{Per-class precision (P), recall (R), and F1 scores.}
\label{tab:perclass}
\small
\begin{tabular}{@{}llccc@{}}
\toprule
\textbf{Model} & \textbf{Class} & \textbf{P $\uparrow$} & \textbf{R $\uparrow$} & \textbf{F1 $\uparrow$} \\
\midrule
\multirow{5}{*}{\shortstack[l]{BiLSTM-\\GRU}}
  & Comprehension & \textbf{0.56} & 0.33 & \textbf{0.42} \\
  & Search        & 0.34 & \textbf{0.63} & \textbf{0.44} \\
  & Speech        & \textbf{0.38} & 0.10 & 0.16 \\
  & Success       & 0.38 & 0.59 & 0.46 \\
  & Timing        & \textbf{0.45} & \textbf{0.38} & \textbf{0.41} \\
\midrule
\multirow{5}{*}{\shortstack[l]{HRNN}}
  & Comprehension & 0.50 & 0.14 & 0.22 \\
  & Search        & \textbf{0.43} & 0.38 & 0.40 \\
  & Speech        & 0.25 & \textbf{0.57} & \textbf{0.35} \\
  & Success       & \textbf{1.00} & \textbf{0.88} & \textbf{0.93} \\
  & Timing        & 0.33 & 0.25 & 0.29 \\
\bottomrule
\end{tabular}
\vspace{2pt}
\end{table}

\begin{table}[t!]
\caption{Task 3 recovery preference prediction results. QLoRA is a fine-tuning method for Mistrial. The base Gemma-9B model produced near-zero valid outputs and is omitted from this table for brevity. See Section \ref{sec:task3_res} for details $^{\dagger\dagger}$.}
\centering
\small
\begin{tabular}{l c c c c c}
\toprule
\textbf{Model} & \textbf{k} & \textbf{Hit@k $\uparrow$} & \textbf{P@k $\uparrow$} & \textbf{R@k $\uparrow$} & \textbf{F1@k $\uparrow$} \\
\midrule
\multirow{5}{*}{Mistral-7B} 
 & 1 & 0.35 & 0.35 & 0.09 & 0.14 \\
 & 2 & 0.62 & \textbf{0.38} & 0.21 & 0.26 \\
 & 3 & 0.65 & 0.32 & 0.26 & 0.28 \\
 & 4 & \textbf{0.68} & 0.30 & 0.30 & 0.29 \\
 & 5 & \textbf{0.68} & 0.29 & \textbf{0.34} & \textbf{0.30} \\
\midrule
\multirow{5}{*}{\shortstack{Mistral-7B \\ QLoRA}} 
 & 1 & 0.38 & 0.38 & 0.10 & 0.16 \\
 & 2 & 0.65 & \textbf{0.39} & 0.22 & 0.27 \\
 & 3 & 0.70 & 0.32 & 0.26 & 0.28 \\
 & 4 & 0.73 & 0.30 & 0.3 & 0.30 \\
 & 5 & \textbf{0.76} & 0.30 & \textbf{0.40} & \textbf{0.32} \\
\midrule
\multirow{5}{*}{\shortstack{Gemma-9B \\ QLoRA}} 
 & 1 & 0.35 & 0.35 & 0.09 & 0.14 \\
 & 2 & 0.62 & \textbf{0.38} & 0.21 & 0.26 \\
 & 3 & \textbf{0.68} & 0.32 & 0.26 & 0.28 \\
 & 4 & \textbf{0.68} & 0.30 & 0.30 & 0.29 \\
 & 5 & \textbf{0.68} & 0.29 & \textbf{0.34} & \textbf{0.30} \\
\bottomrule
\\
\end{tabular}

\vspace{4pt}
{$^{\dagger\dagger}$\,See Table~\ref{tab:window-classification} footnote. }

\label{tab:task3_results}
\end{table}

\subsection{Task 3: Recovery Preference Prediction}

\textbf{Base Model Comparison.} \label{sec:task3_res} Table~\ref{tab:task3_results} shows Hit@$k$, Precision@$k$, Recall@$k$, and F1@$k$ for $k=\{1, 2, 3, 4, 5\}$. 
We evaluate both base instruction models and domain-adapted variants fine-tuned using QLoRA \cite{qlora}: Mistral-7B (base), Mistral-7B fine-tuned, Gemma-9B (base), and Gemma-9B fine-tuned{\cite{jiang2023mistral7b, GoogleGemma2024}}.

Among base models, Mistral-7B outperforms Gemma-9B across all Top-$k$ metrics, highlighting the advantage of instruction-tuned LLMs for structured output generation tasks. Gemma-9B’s base performance is near 0 across all Top-$k$ metrics (Hit@$k$, Precision@$k$, Recall@$k$, F1@$k$), likely because it is not instruction-tuned. This indicates that even large generalist LLMs struggle to generate valid structured recovery strategy sets without task-specific adaptation. 

\textbf{Effect of QLoRA Fine-Tuning.} Across both base models, QLoRA fine-tuning consistently improves performance for all Top-$k$ metrics. For example, Mistral-7B shows a modest Top-1 Hit increase from 0.3514 to 0.3784 (+7.8\%) and an F1@5 increase from 0.3017 to 0.3207 (+6.2\%). Gemma-9B, which performed at chance levels in its base form, achieves comparable performance to Mistral-7B after QLoRA fine-tuning, demonstrating that domain adaptation via parameter-efficient fine-tuning is critical for aligning LLMs with task-specific recovery prediction.

\textbf{Trends Across k.} Improvements are most pronounced at higher $k$ values, particularly for F1, recall, and Hit metrics. This indicates that fine-tuned models better capture the set of acceptable user-preferred recovery strategies, rather than predicting the single most likely action. The plateau observed in Mistral-7B and Gemma-9B QLoRA at $k=4$–$5$ suggests that more candidate strategies are increasingly difficult to rank accurately, reflecting inherent ambiguity in user preferences.

\section{Conclusion}

We presented REPAIR-Bench, the first end-to-end benchmark for multimodal failure detection and recovery prediction in healthcare HRI. Built on the RFM-HRI dataset \cite{rfm-hri}, the benchmark introduces failure time annotations, two baseline model pipelines (an HRNN for detection/classification and an SLM for recovery), and a standardized evaluation protocol. By integrating detection, classification, and recovery into a single framework, REPAIR-Bench enables systematic study of how robots can not only recognize failures but also recommend appropriate, user-preferred repair strategies.
Our results show that fine-tuned models are better equipped to recommend user-preferred recovery actions, an essential capability in safety-critical healthcare settings where multiple interventions may be appropriate. 

Some limitations include the relatively small number of recovery prediction samples (184), which may limit generalizability, and the use of a predefined set of recovery strategies presented to users. Future work will expand the dataset, incorporating richer multimodal signals such as facial expressions, gestures, and gaze alongside textual cues, and extend the benchmark to additional HRI domains to further validate robustness and real-world applicability.

\section*{Acknowledgment}

% ADD BACK AFTER PAPER DECISION
% \textit{Grant Acknowledgment.} \thanks{This material was supported by the National Science Foundation under Grant No. IIS-2423127.}

We would like to acknowledge the use of generative AI tools, including ChatGPT and Gemini, in the preparation of this manuscript. Specifically, AI was employed to review the text and provide suggestions for grammar, clarity, and readability throughout the paper, all outputs were carefully reviewed and validated by the authors. All initial drafts, analyses, interpretations, and conclusions were authored entirely by the human authors. No AI generated content was used in figures, tables, or scientific analysis. All intellectual responsibility for the work remains with the authors.

\bibliographystyle{IEEEtran}
\bibliography{references.bib}

\end{document}